\documentclass{article}

\usepackage{arxiv}

\usepackage[utf8]{inputenc} % allow utf-8 input
\usepackage[T1]{fontenc}    % use 8-bit T1 fonts
\usepackage{hyperref}       % hyperlinks
\usepackage{url}            % simple URL typesetting
\usepackage{amsmath,amssymb,amsfonts,bm}
\usepackage{graphicx}
\graphicspath{ {./images/} }
\usepackage{subfigure}

\newcommand{\titletext}{Control of Walking Assist Exoskeleton with Time-delay Based on the Prediction of Plantar Force}
\title{\titletext}

\author{
 Ming Ding \\
  Institutes of Innovation for Future Society \\
  Nagoya Univeristy \\
  Nagoya, Aichi 464-8601, Japan \\
  \texttt{dingming@nagoya-u.jp} \\
   \And
 Mikihisa Nagashima\\
  Division of Information Science\\
  Nara Institute of Science and Technology\\
  Ikoma, Nara 630-0192 Japan \\
  \texttt{nagashima.mikihisa.ni8@is.naist.jp} \\
  \And
 Sung-Gwi Cho \\
  Division of Information Science\\
  Nara Institute of Science and Technology\\
  Ikoma, Nara 630-0192 Japan \\
  \texttt{cho@is.naist.jp} \\
  \And
 Jun Takamatsu \\
  Division of Information Science\\
  Nara Institute of Science and Technology\\
  Ikoma, Nara 630-0192 Japan \\
  \texttt{j-taka@is.naist.jp} \\
  \And
 Tsukasa Ogasawara \\
  Division of Information Science\\
  Nara Institute of Science and Technology\\
  Ikoma, Nara 630-0192 Japan \\
  \texttt{ogasawar@is.naist.jp} \\
}

\begin{document}

\rhead{\titletext\ - \today}

\twocolumn[
\maketitle

\begin{abstract}
Many kinds of lower-limb exoskeletons were developed for walking assistance.
However, when controlling these exoskeletons, time-delay due to the computation time and the communication delays is still a general problem.
In this research, we propose a novel method to prevent the time-delay when controlling a walking assist exoskeleton by predicting the future plantar force and walking status.
By using Long Short-Term Memory and a fully-connected network, the plantar force can be predicted using only data measured by inertial measurement unit sensors, not only during the walking period but also at the start and end of walking.
From the predicted plantar force, the walking status and the desired assistance timing can also be determined.
By considering the time-delay and sending the control commands beforehand, the exoskeleton can be moved precisely on the desired assistance timing.
In experiments, the prediction accuracy of the plantar force and the assistance timing are confirmed.
The performance of the proposed method is also evaluated by using the trained model to control the exoskeleton.
\end{abstract}

% keywords can be removed
\keywords{Walking exoskeleton \and Motion Prediction \and Machine Learning \and Plantar Force.}

\vspace{\baselineskip}
]

\section{Introduction}
\label{sec:intro}

Walking assist exoskeleton is a device to help the wearer walk with his/her own legs for power amplification or assistance.
Such exoskeletons have been widely researched for rehabilitation and power assist~\cite{Dollar_2008,Hayashi_2005}.
Today, many exoskeletons are controlled based on the measurement of human motion and force~\cite{Bleex2006,wear2005}.
However, there is still a problem of time-delay due to the required communication, computing, and response time that occur between measuring the human motion and producing an appropriate assisting force.
The time-delay can be reduced by predicting the future human motion from the measured walking status to realize on-time and high-fidelity control.

Some biosignals such as electromyogram (EMG) and electroencephalogram (EEG) can be measured before the movement of muscles~\cite{lee2002}.
By predicting the human motion from such biosignals, an exoskeletons can be controlled in accordance with the wearer's motion intention~\cite{kiguchi2007}.
However, the electrodes for measuring the biosignals should be attached directly to the skin,
and the measured signal may be affected by the deformation of skin, sweat, the location and orientation of electrodes.
It is difficult to get such biosignals stably and reproducibly.
Some other methods predict the walking status by estimating the walking cycle using the inertial measurement unit (IMU) sensor~\cite{Stolyarov_2018}.
During the stable walking (i.e., walking in a regular rhythm), such methods are effective to predict the walking status and assistance timing.
However, it is difficult to get the assistance timing at the start and end of a walking.

\begin{figure*}[t]
    \centering
    \includegraphics[width=0.85\linewidth]{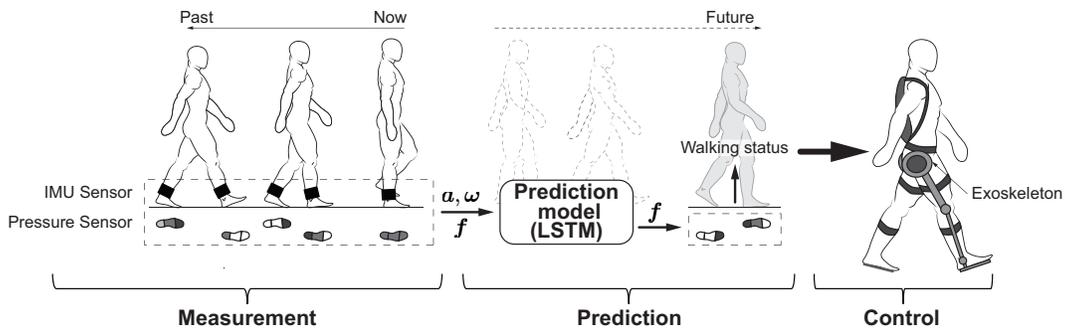}
    \caption{Prediction of plantar force for control of a walking assist exoskeleton}
    \label{fig:concept}
\end{figure*}

In many studies on rehabilitation and sports, the plantar force has been widely used for gait analysis~\cite{Razak2012FootPP}.
The change in the plantar force always happens when moving the human body through movement of the center of gravity (COG) and the grounding status of the feet.
By analyzing the change of plantar force, the walking phases can be detected.
However, for measuring the plantar force in long-term walking, a pressure sensor with high accuracy and high duration is necessary.
The cost and size of such pressure sensor is still a bottleneck when using an exoskeleton with different shoes.

Today, with the development of deep learning technology, human motion prediction has been studied based on the deep recurrent neural networks~\cite{fragkiadaki2015,Jain2016}.
These works showed good performance when predicting the future motion signals (i.e., joint positions and angles) from the measured signals using motion capture systems or outside cameras.
However, for using an exoskeleton in a wide space, it is still necessary to measure and predict the motion from the wearable sensors.

In this research, as shown in Fig.~\ref{fig:concept}, we propose a novel method to prevent the time-delay when controlling a walking assist exoskeleton by predicting the future plantar force from wearable IMU sensors.
From the predicted plantar force, the walking status is estimated to detect the timing of walking assistant.
Based on the Long Short-term Memory (LSTM), a network is constructed to learn the relationship between the measured motion data from IMU sensors and the future plantar force data from pressure sensors.
By considering the time-delay and sending the control command beforehand, on-time walking assistance is achieved.
In experiments, we evaluate the performance both quantitatively and qualitatively by analyzing the prediction accuracy and controlling an exoskeleton in real-time.

% In this work, the following two primary contributions have been made.
% (1) We proposed and verified a method to predict the plantar forces from the signal measured only by IMU sensors.
% (2) We verified that the time-delay when controlling a walking assist exoskeleton could be prevented based on the predicted plantar force and walking status.
 
\section{Proposed method}
\label{sec:method}

As shown in Fig.~\ref{fig:concept}, our method consists of three stages, discussed in this section.
First, we measure the walking motion data from the worn IMU sensors (i.e., the acceleration $\bm{a}$ and the angular velocity $\bm{\omega}$).
For training the prediction model, the plantar force is also measured by putting a pair of insole-type pressure sensors in the wearer's shoes.
Then, a model is created and trained based on Long short-term memory (LSTM) method~\cite{Hochreiter1997}.
Using the trained model, the plantar force can be predicted from the measured past motion data.
Based on the predicted plantar force, the walking phase is estimated.
Finally, based on the estimated future walking phase, we control an exoskeleton considering the time delay of the device. 
Note that the measurement of plantar force is only needed for training the model.
When controlling the exoskeleton in real-time, only the motion data (i.e., the signals from the IMU sensors) needs to be measured.

\subsection{Measurement of motion data}

In this research, due to wearability and usability, wireless IMU sensors are used to measure the walking motion.
An IMU sensor is generally composed of accelerometer, gyroscope, and sometimes magnetometer, which can measure the acceleration, the angular velocity, and sometimes the direction of the magnetic field.
It has been widely used for the measurement of human motion (e.g., Xsens, Perception Neuron).
To get the motion from the IMU sensors, many calculation methods have been proposed such as using Kalman or Complementary filter~\cite{Filippeschi2017}.
However, because of the integral calculation and noise when calculating the motion, estimation errors and a time-delay always exist.
In this research, we directly use the measured signals (acceleration and angular velocity) as the motion signals without any processing (e.g., to calculate the position and rotation angle), which can save the processing time and reduce the error from the pipe-line calculation.
The measurement noise also need not to be filtered.

However, when wearing an exoskeleton, the movement of the limbs that fixed to the device are controlled by the device.
The motion of these fixed and directly-supported limbs depends on the device and does not react to the intention of the wearer. 
Therefore, to predict the future walking status, we should measure the motion of the limbs not controlled by the exoskeleton.
For example, if the exoskeleton has motors to only assist the movement of the thighs, the IMU sensors should be set up on lower legs or upper body.

On the other hand, we predict the plantar force from the measured motion signals (i.e., the signals from IMU sensor).
The plantar is required to estimate the walking status and detect the assistance timing.
In this research, we put two insole-type pressure sensors in the shoes, which can measure the force between human feet and the ground.
In each pressure sensor, multiple sensor cells are embedded for detect the change in pressure and determine the movement of the human body, such as the Center of Pressure (COP) or the Center of Gravity (COG).
Note that other types of pressure sensors, such as using 3-axis or 6-axis force sensors, could be used if the sensors can measure the data related to the movement of the human body.

\subsection{Prediction of plantar force and walking phase}

From the measured motion data, we created a deep-learning-based model to predict the future plantar force.
The predicted force is then used to estimate the future walking phase.

\subsubsection*{Plantar force:}

If the current time is $t$, the plantar force $\bm{\hat{\bm{y}}_{t+s}}$ at the future time $t+s$ is predicted from the past signals $[{\bm{x}_{t}, \bm{x}_{t-1}, \ldots, \bm{x}_{t-n+1}}]$ measured by the IMU sensors as follows:
\begin{equation}
\label{eq:yfx}
\hat{\bm{y}}_{t+s} = f(\bm{x}_{t}, \bm{x}_{t-1}, \ldots, \bm{x}_{t-n+1}),
\end{equation}
where $s$ is the prediction time.
$\bm{x}_{t}, \ldots, \bm{x}_{t-n+1}$ are the measured data from time $t$ to time $t-n+1$.
If we wear $m$ IMU sensors on the human body, $\bm{x}$ is a vector with $ 6 \times m$ elements (acceleration $\bm{a}$ and angular velocity $\bm{\omega}$), $\bm{x} = [\bm{a}_1, \bm{\omega}_1, \ldots, \bm{a}_m, \bm{\omega}_m]^T \in \bm{R}^{\{6 \times m\} \times 1}$.
% \begin{equation}
% \bm{x} = [\bm{a}_1, \bm{\omega}_1, \ldots, \bm{a}_m, \bm{\omega}_m]^T
% \end{equation}
If there were $k$ cells in each plantar sensor, the predicted plantar force is a vector with $k\times 2$ elements, $\hat{\bm{y}} \in \bm{R}^{\{k \times 2\} \times 1}$.

\begin{figure}[t]
    \centering
    \includegraphics[width=1.0\columnwidth]{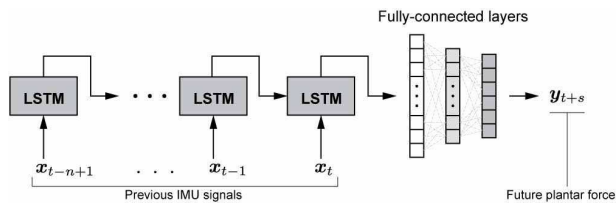}
    \caption{Architecture of the prediction neural network}
    \label{fig:network}
\end{figure}

As shown in Fig.~\ref{fig:network}, a deep learning model using LSTM and fully-connected layers is designed to define the prediction function \eqref{eq:yfx}.
The input of the LSTM cell in every time $t$ is the motion data $\bm{x}_t$ (i.e., the signal measured from the IMU sensors).
We set the number of hidden states of each LSTM cell to 32.
% We set 32 hidden states.
When inputting $n$ flames of past measured signals, the last output of the hidden state is input to a 3-layer fully-connected network with ReLU activation.
The size of the input layer of this network is the same as the number of the hidden states (i.e., 32 neurons).
The size of the output layer is the same as the number of the cells in two plantar sensors (i.e., $k \times 2$ neurons).
One hidden layer with 16 neurons is used to obtain the non-linear relationship between the phase and the predicted values.

For training the network, we take $n$ frames of the measured past signals of the IMU sensors as input.
The measured plantar force after $s$ frames is used as the target output.
To use the trained network, we input the measured past $n$-frame signals from the IMU sensors, and the network will output the predicted plantar force of the future $s$ frame.
When using this method, only the measured past signals should be used.
Therefore, the prediction can also be used to control the exoskeleton in real-time.

\subsubsection*{Walking phase:}

\begin{figure}[t]
    \centering
    \includegraphics[width=1.0\columnwidth]{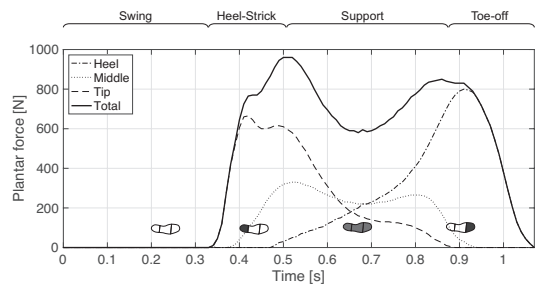}
    \caption{The platar force of left foot during one walking cycle}
    \label{fig:onecycle}
\end{figure}

The walking phase is estimated based on the predicted plantar force.
Result from many other studies shows that the plantar force is related to the walking phases (i.e., swing phase and stance phase).
The detailed walking stage (i.e., heel-strike, support, toe-off, swing) can also be estimated.

Fig.~\ref{fig:onecycle} shows the change in the plantar force of the left foot during one walking cycle from a heel-strike to the next heel-strike.
The plantar force starts on the heel $f_h$ and increases to the first peak during the heel-strike stage.
Then, during the support stage, the plantar force wqualizes across all pressure cells on the foot.
Before the foot swing, the plantar force on the tip $f_t$ increases to the second peak for kicking the ground.
When foot start the swing, the plantar force is zero, $f=0$.
By detecting the changing points for each plantar force ($f_h$, $f_m$, and $f_t$), the walking phases can also be estimated.

\subsection{Control of exoskeleton}

Finally, based on the predicted walking phase, we detect the assistance timing needed.
When controlling the actuators to assist the human motion, most exoskeletons have a time-delay ($t_d$),  mainly including the measurement time of the signal ($t_{dm}$), the computation time ($t_{dc}$), and the response time of the actuators ($t_{dr}$),
\begin{equation}
    t_d = t_{dm} + t_{dc} + t_{dr}.
\end{equation}
To prevent such time-delay, the walking phase in $t_d$ future should be estimated.

In this research, the assistance timing is detected by estimating the change in the plantar force.
Therefore, the prediction time $s$ should be a little larger than the time-delay $t_d$ ($t_d < s$).
For example, to increase the kicking force applied to move the body forward, an assistance force should be applied when starting to toe-off phase marked by the change of $f_t$.
To reduce the strike on the knee and heel, the assistance force should be applied during the heel-strike phase marked by the change in $f_h$.
Note that other methods, such as using matching and learning methods, could be used to detect the walking stages.
After detecting the assistance timing, a control command is sent $t_{dr}$ before the motors exactly on the necessary timing.

\section{Experiments}
\label{sec:exps}

To confirm the proposed method, two types of experiments were performed.
In the first experiment, we quantitatively verified the prediction accuracy of the plantar force and the assistance timing.
In the second experiment, we evaluated the effectiveness when controlling the exoskeleton by measuring the motion and predicting the walking phase in real-time.
The experimental protocol of this study was approved by the research ethics board of the Nara Institute of Science and Technology.

\subsection{Experimental system}

\subsubsection{Sensors}

\begin{figure}[t]
    \centering
    \subfigure[IMU sensor]{
    \includegraphics[height=0.4\columnwidth]{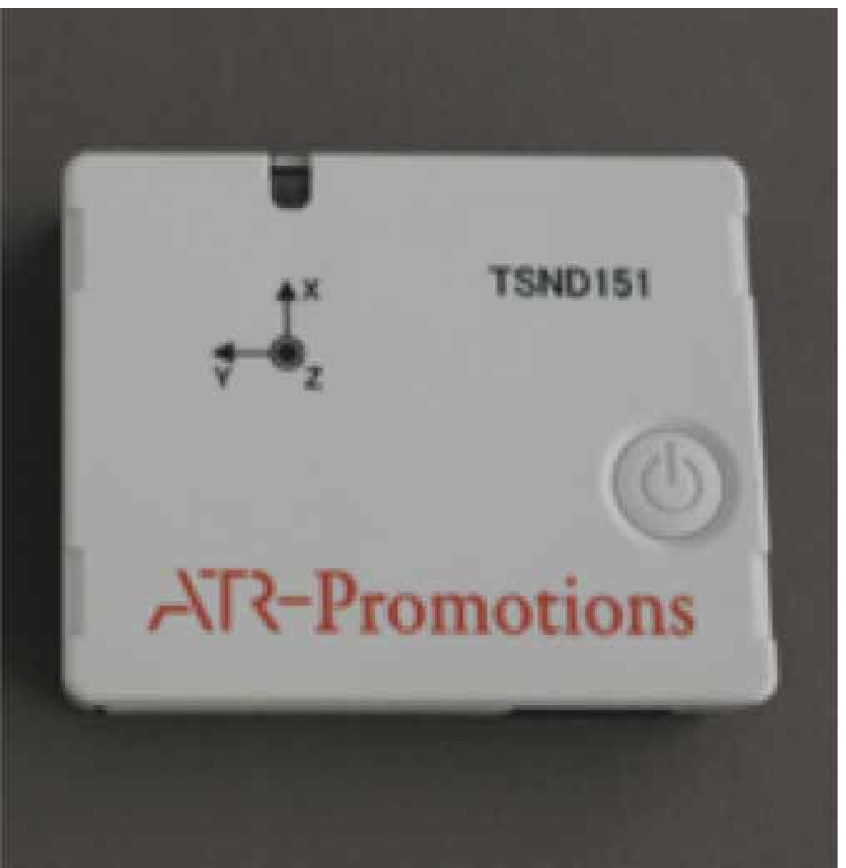}
    \label{fig:IMU}
    }
    \subfigure[Plantar sensor]{
    \includegraphics[height=0.4\columnwidth]{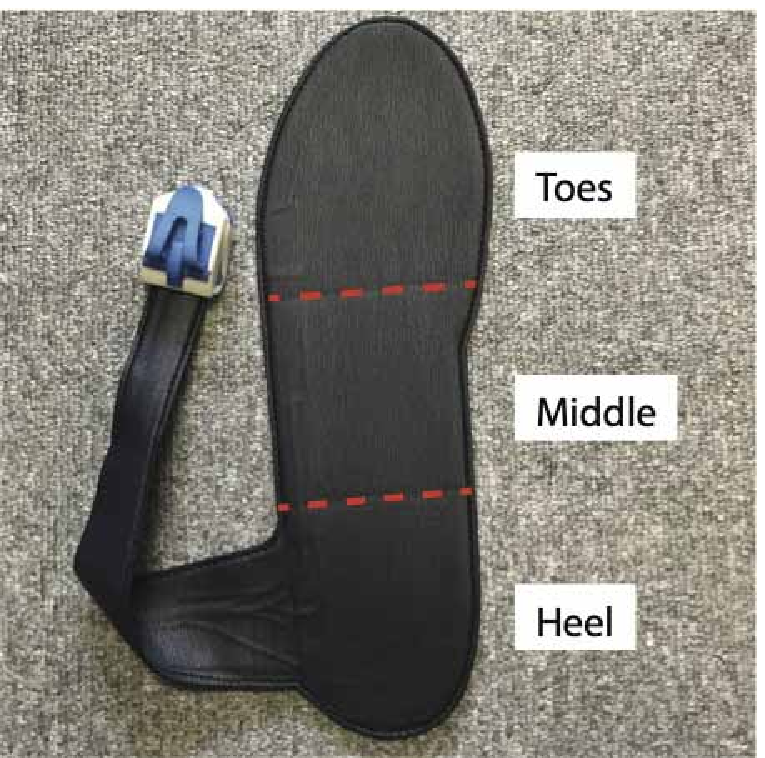}
    \label{fig:loadsol}
    }
    \caption{The wearable sensors used}
    \label{fig:sensors}
\end{figure}

In the experiments, wearable plantar sensor and IMU sensor were used since they can applied easily even while wearing a lower-limb exoskeleton.
Fig.~\ref{fig:IMU} shows the wireless IMU sensor (TSND151, ATR-Promotions) used in this research.
This sensor can measure acceleration and angular velocity simultaneously and transmit the data at up to 1000Hz via Bluetooth.
Fig.~\ref{fig:loadsol} shows the wireless plantar sensor (Loadsol, Novel) we used.
It is a soft insole-type sensor and can monitor the normal force between the plantar side of the foot and the shoe.
There are three subareas (i.e., three cells) in the sensor to measure the change of the pressure on each foot.
The normal contact forces from forefoot (toes), midfoot (middle), and hindfoot (heel) can be measured with an accuracy of 10N and a data transmission speed of up to 200Hz via Bluetooth.

\begin{figure}[t]
    \centering
    \includegraphics[height=0.6\columnwidth]{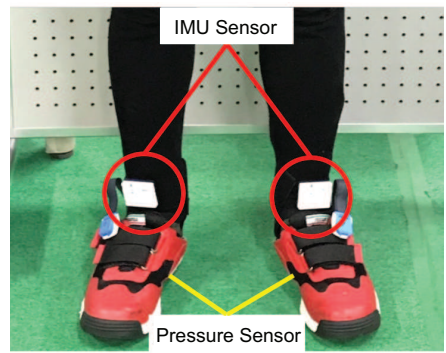}
    \caption{Setup of the wearable sensors}
    \label{fig:IMU_place}
\end{figure}

As shown in Fig.~\ref{fig:IMU_place}, we placed the IMU sensors on the lower legs near the ankles and put the plantar sensors in the two shoes.
The data measured from the four sensors was resampled to 100Hz.
 
\subsubsection{Walking assist exoskeleton}

\begin{figure}[t]
    \centering
    \subfigure[Front view]{
    \includegraphics[height=0.60\columnwidth]{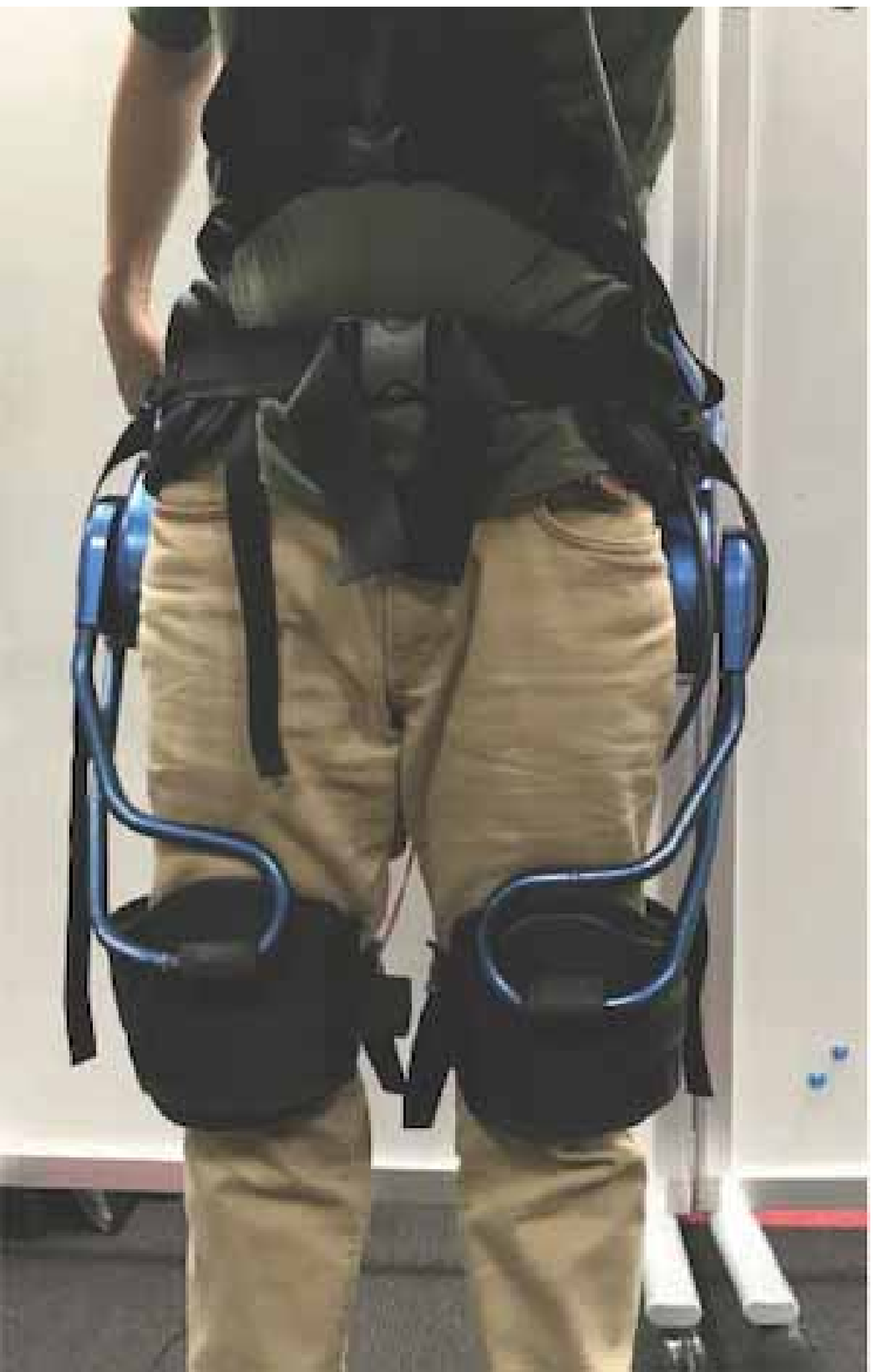}
    }
    \subfigure[Side view]{
    \includegraphics[height=0.60\columnwidth]{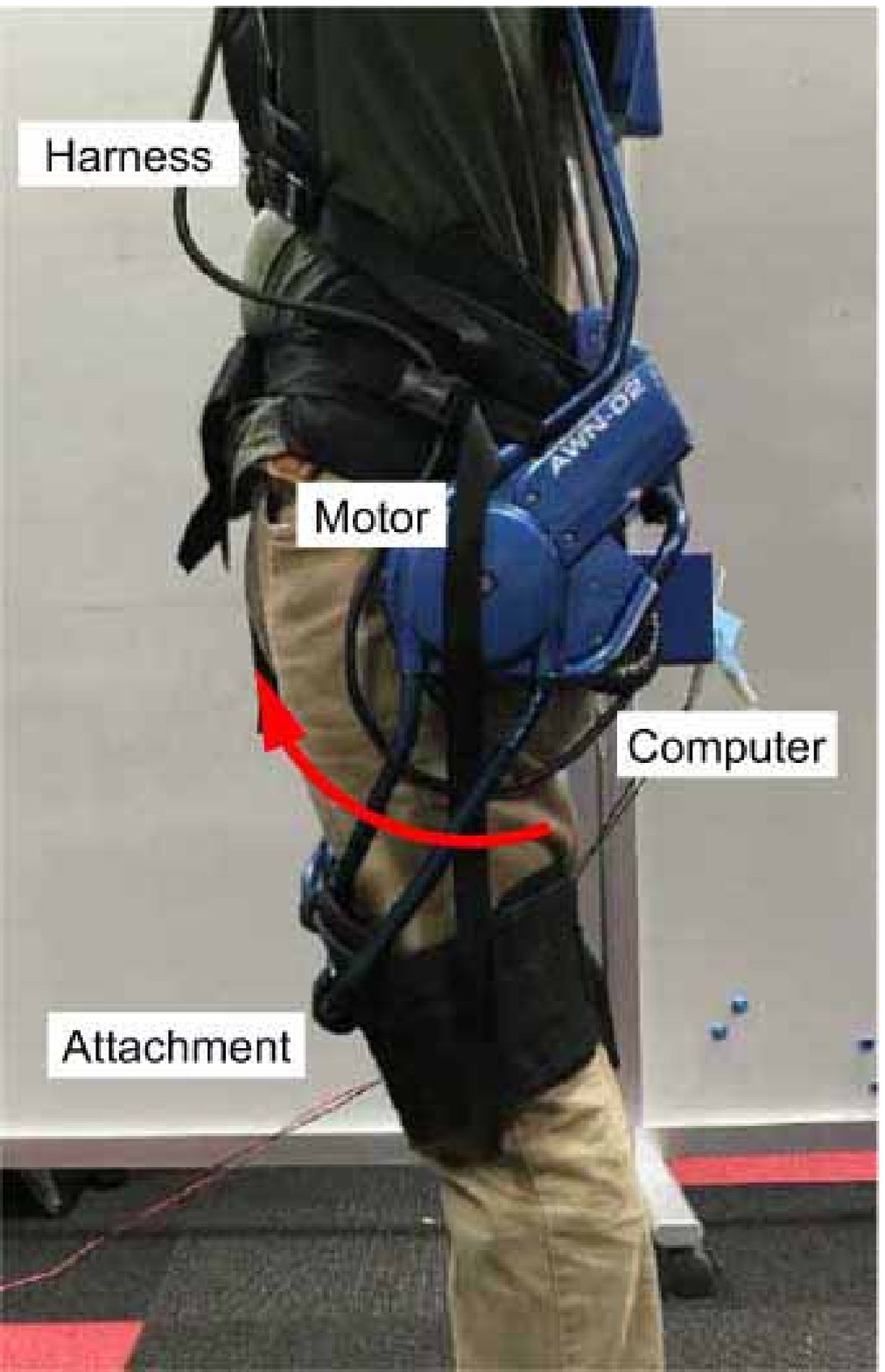}
    }
    \caption{The used lower-limb exoskeleton (AWN-02, ATOUN)}
    \label{fig:AWN-02}
\end{figure}

We tested the proposed method by implementing it with a lower-limb exoskeleton device (AWN-02, Atoun inc.).
As shown in Fig.~\ref{fig:AWN-02}, with this device, there are two motors on each side of the hips.
The motors are connected to the thighs near the knees using the rigid links and velcro-tape attachments.
By rotating the motors, a force can be applied to rotate the thigh forward.
The exoskeleton can move the leg and foot up and reduce the force needed to lift and swing the legs.
This exoskeleton only controls the movement of the thighs.
The lower legs where we placed the IMU sensors are free from the exoskeleton, which is suitable for using our method to determine the wearer's intention for motion.

We assembled a small single board computer (Raspberry Pi 3B) in the exoskeleton.
This computer can measure the signals from the IMU sensors through the Bluetooth communications.
The trained network can be copied to this computer to predict the walking phases in real-time from the measured IMU signals.
This computer is also used to control the exoskeleton by sending control signals to the control board through serial communication.

\begin{table}[t]
    \caption{Required computation time of prediction when using the trained model in small computer (Raspberry PI 3B)}
    \label{tab:prediction_time}
    \centering
    \begin{tabular}{c|c|c}
        \hline
        Used time [s] & Used frames ($n$) & Computation time $t_{dc}$ [s] \\
        \hline
        0.01 & 1 & 0.010  \\
        0.1 & 10 & 0.018  \\
        0.2 & 20 & 0.024  \\
        0.3 & 30 & 0.035  \\
        0.4 & 40 & 0.042  \\
        \hline
    \end{tabular}
\end{table}

By measuring the time difference between motion signal and motor power, we calculated the time-delay between signal measurement and motor response is about 0.1s (i.e., $t_{dm}+t_{dr} \approx 0.1 \text{s}$).
The time-delay of computation (i.e., the time for the prediction using the trained network) depends on the number of the used frames of measured past signals $n$.
Table~\ref{tab:prediction_time} shows the computation time for different $n$.
For example, if we use 0.2s past signals, the prediction time $t_{dc}$ would be about 0.024s, so the total time-delay would be about 0.124s, $t_d = (t_{dm}+t_{dr})+t_{dc} \approx 0.124 \text{s}$.
The prediction time $s$ should be larger than it (i.e., $s>124$).

\subsection{Experiment 1: Prediction of Plantar force}

In this experiment, we measured the walking data from nine healthy subjects (Males, $24.4\pm1.6$ y/o, $72.7\pm11.3$kg, $172.2\pm5.2$cm) when not wearing the exoskeleton.
This was to verify whether the proposed method accurately predicted the plantar force and the assisting timing.

\begin{figure}[t]
    \centering
    \includegraphics[width=1.0\columnwidth]{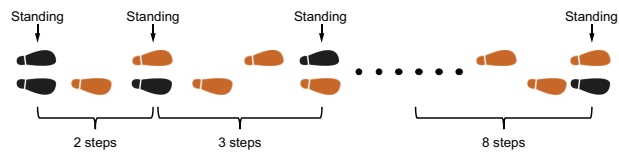}
    \caption{An example of the steps in one walking trail}
    \label{fig:walkingtrail}
\end{figure}

A japanese textbook about body dynamics described that human's walking motion would become stable after about four steps.
To test both stable and unstable walking,  as shown in Fig.~\ref{fig:walkingtrail}, we asked the subjects to walk from two steps to eight steps in every trail.
Between two sets of walking, the subjects should pause and stand a moment before starting the next walking.
Every subject performed such walking 8 trails and also performed 2 trails walking with random steps and pauses as desired.
Totally, from every subject, we measured the walking data of 10 trails.
In five of the trials, as shown in Fig.~\ref{fig:walkingtrail}, they started walking with the right leg.
In the other five trails, they started walking with the left leg.

The time of one walking trail was about 30 seconds (30,000 frames).
For every subject, we collected about five minutes of walking data using the IMU sensors and the plantar force sensors.
The paired input data (motion data from IMU sensors) and the desired output data (plantar force from Pressure sensors) was created for every 5 frames (0.05s).
We trained and tested the model for each subject individually.
The data of eight trails were used to train the proposed prediction model (Fig.~\ref{fig:network}).
The remaining 2 trials (including one patterned walking and one random walking) were used for the test.

To confirm the performance of the proposed method, we tested the prediction accuracy with different lengths of the measured past IMU signals: 0.01s ($n=1$), 0.05s ($n=5$), 0.1s ($n=10$), 0.2s ($n=20$), 0.5s ($n=50$), and 1s ($n=100$).
We also tested the accuracy at predicting the plantar force with different future time: 0.01s ($s=1$), 0.1s ($s=10$), 0.2s ($s=20$), 0.5s ($s=50$), 1s ($s=100$), 1.5s ($s=150$), 2s ($s=200$), and 5s ($s=50$).
The prediction error is defined as the difference between the predicted and measured plantar force.

\subsubsection*{Results}

\begin{figure}[t]
    \centering
    \subfigure[Test $n$ when $s=20$]{
        \includegraphics[width=0.47\columnwidth]{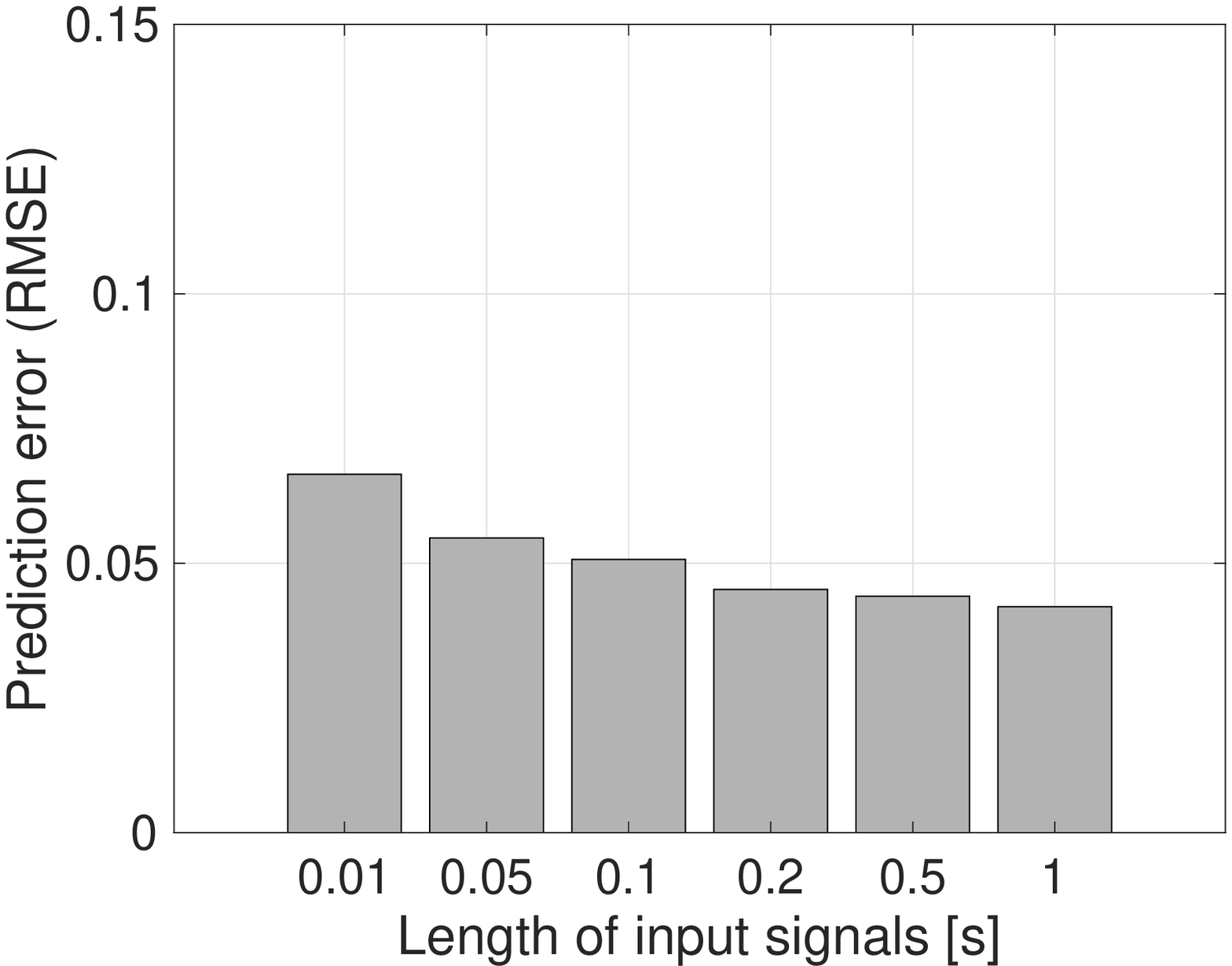}
        \label{fig:test_np}
    }
    \subfigure[Test $s$ when $n=20$]{
        \includegraphics[width=0.47\columnwidth]{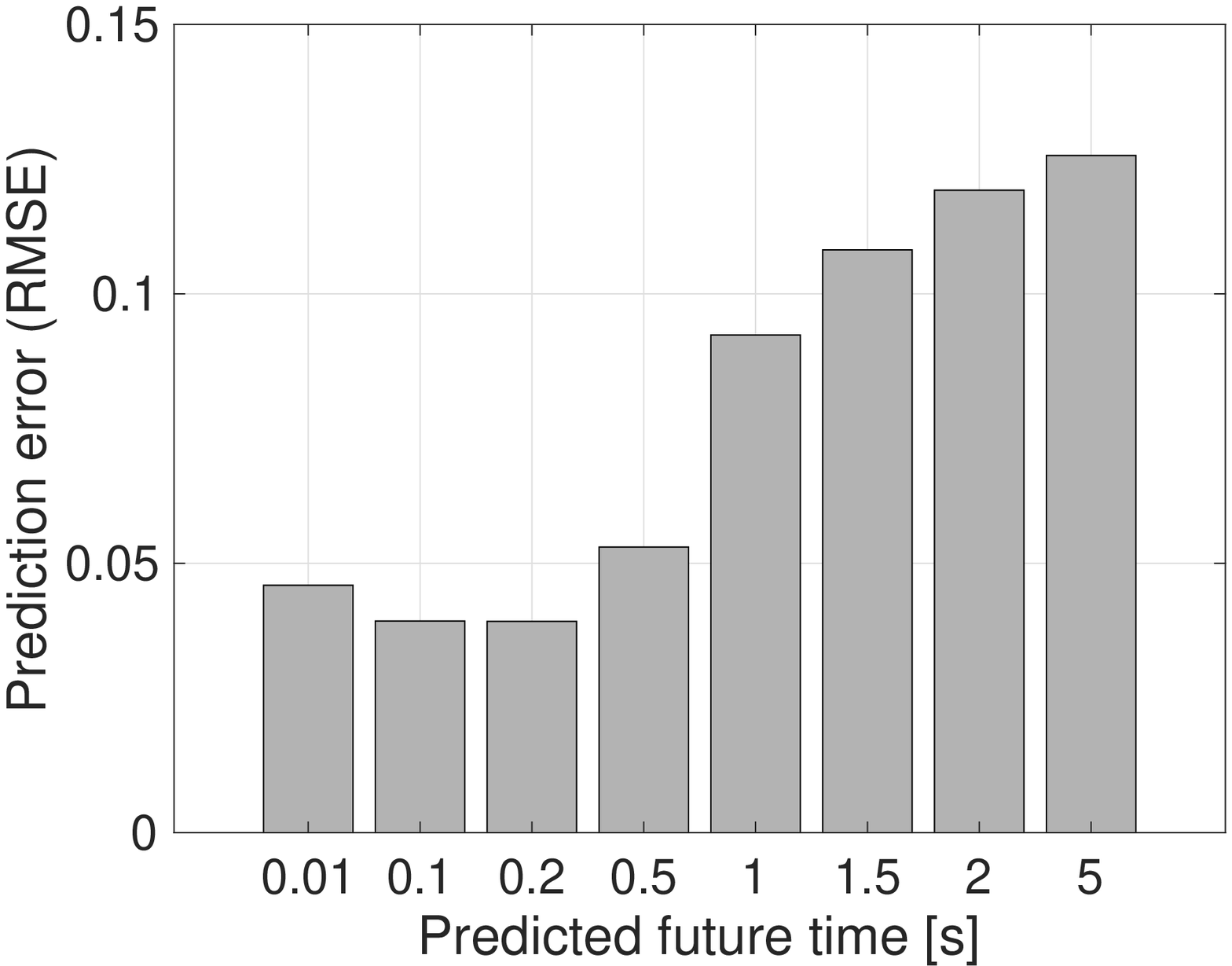}
        \label{fig:test_nf}
    }
    \caption{Prediction errors for different length of input signals and future time}
    \label{fig:test_inout_num}
\end{figure}

Fig.~\ref{fig:test_inout_num} shows the average prediction error for all the testing data from all the subjects using different length of input IMU signals and predicting different future plantar forces.
As shown in Fig.~\ref{fig:test_np}, using more past IMU signals reduced the prediction error from about 6.7\% to 4.2\%.
However, even using more signals, the change was small.
Considering the computing performance of the small computer (Raspberry PI 3B) used in this research, 0.2s IMU signals ($n=20$) were used for the prediction.
Fig.~\ref{fig:test_nf} shows the average prediction error for different futures when inputting the same measured IMU signals (0.2s, $n=20$).
With an increase in the prediction time $s$, the prediction errors also become larger (from about 4.6\% to 12.6\%).
To achieve high-accuracy predictions, the prediction time should be as shorter as possible.
As described above, using these sensors and exoskeleton, the prediction time $s$ should be larger than 0.124s.
Therefore, $s = 20$ was used in the following experiments.

\begin{figure*}[t]
    \centering
    \includegraphics[width=0.8\linewidth]{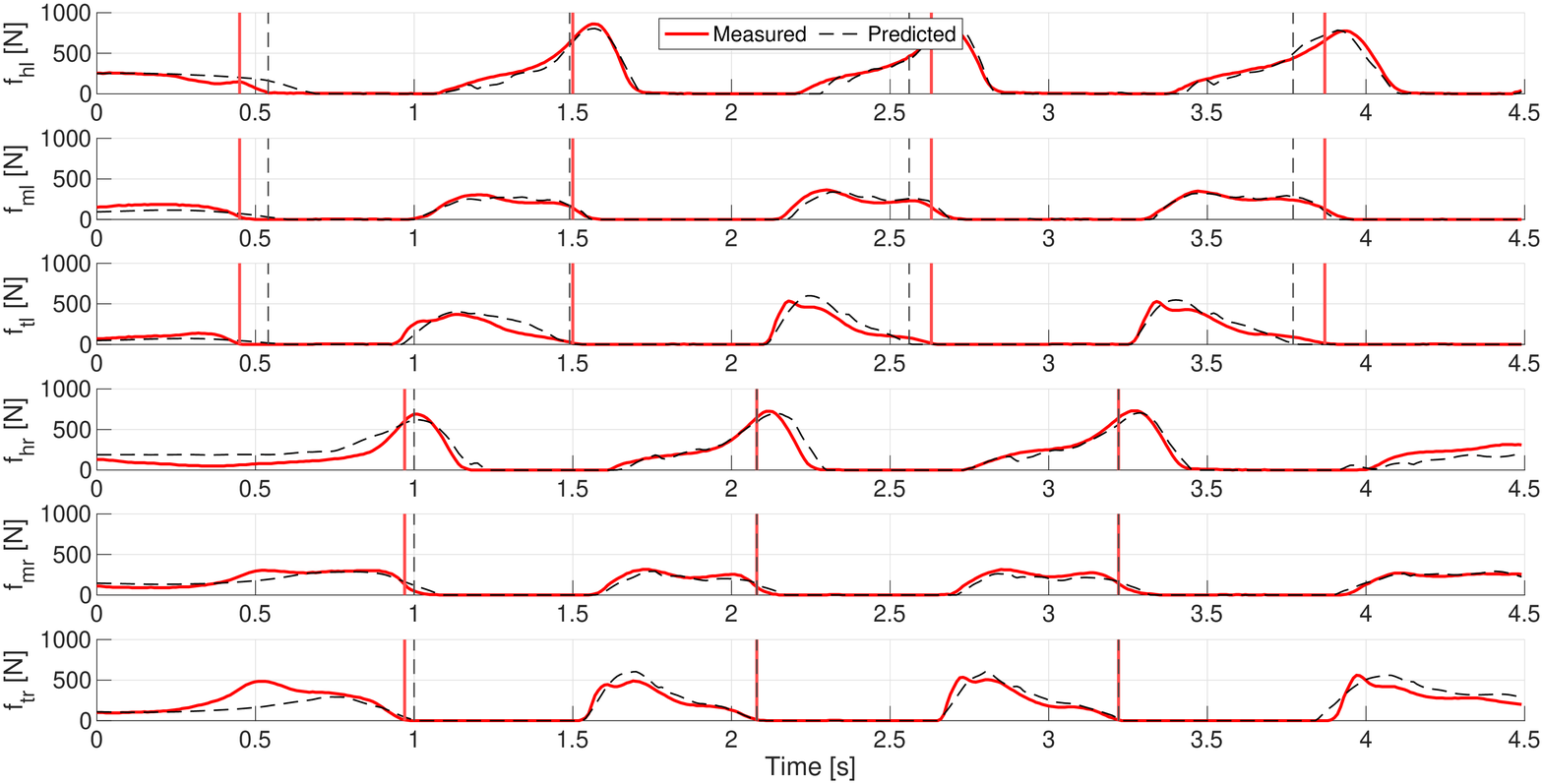}
    \caption{The prediction results of a walking without wearing exoskeleton}
    \label{fig:pred_nodev}
\end{figure*}

Fig.~\ref{fig:pred_nodev} shows the prediction result for a six-step walking by a subject in a trail.
After 0.2s ($n=20$), the plantar force was predicted using the measured IMU signals for the past 0.2s ($s=20$).
The graphs show the plantar force of each cell in order: left heel $f_{hl}$, left middle $f_{ml}$, left toe $f_{tl}$, right heel $f_{hr}$, right middle $f_{mr}$, and right toe $f_{tr}$.
The red solid lines show the predicted force and the blue dashed lines show the measured values using the pressure sensors.
In this six-step walking, the subject started walking with the left leg and stopped with the right leg.
The predicted values do not deviate largely from the actual measured values.
Even for the start and end of walking, the predicted plantar force shows the same change as the measured value.

\begin{figure}[t]
    \centering
    \includegraphics[width=0.95\columnwidth]{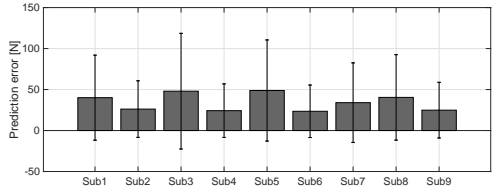}
    \caption{Prediction errors of each subject without wearing exoskeleton}
    \label{fig:prederr_nodev}
\end{figure}

Fig.~\ref{fig:prederr_nodev} shows the average prediction errors of each subject for all frames and all ten trails.
All prediction errors are less then 50N (23.5\textasciitilde48.8N), which is about 7.5\% of the average body weight of subjects.
The average prediction error for all trails and all subjects is 33.9N.

\begin{figure}[t]
    \centering
    \subfigure[Assistance timing while walking]{
    \includegraphics[width=0.95\columnwidth]{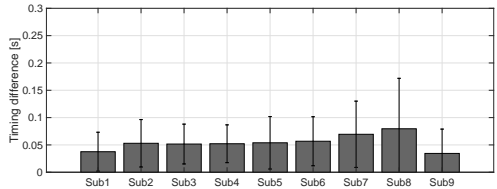}
    \label{fig:pred_tmg_err_nodev_walking}
    }
    \subfigure[Assistance timing when starting]{
    \includegraphics[width=0.95\columnwidth]{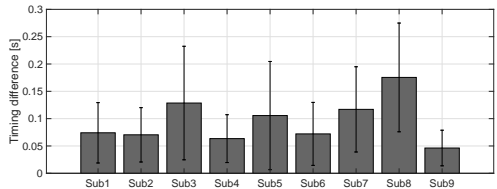}
    \label{fig:pred_tmg_err_nodev_start}
    }
    \caption{Difference in the assistance timings without wearing exoskeleton}
    \label{fig:pred_tmg_err_nodev}
\end{figure}

Based on the predicted plantar force, we detected the assistance timing needed to support walking during the swing stage.
The assistance timing is planned for when the plantar force on toe is smaller than 50N ($f_t \le 50N$).
This level was set by trial and error, considering the measurement resolution of the pressure sensor and the prediction error.
Note that when using other exoskeleton to support walking in other stages, the assistance timing could also be detected in other thresolds or methods.
The vertical lines in Fig.~\ref{fig:pred_nodev} show the detected assistance timings based on the measured and predicted plantar force.
The red vertical lines show the predicted assistance timing while the black dash lines show the assistance timing calculated from measured data.
The assistance timings calculated from the predicted and measured data are very close to each other.
Fig.~\ref{fig:pred_tmg_err_nodev} shows the differences in the assistance timings of all subjects.
During the walks, the differences in the assistance timings of all subjects are less than 0.079s, and the average difference is 0.055s.
When starting the walks, the assisting timing can also be detected.
The difference of all subjects are smaller than 0.175s, and the average difference for all subjects is 0.093s.
It is larger than the difference during the walking, but still small enough for detecting the walking stages.

\subsection{Control of exoskeleton}

In this experiment, the subjects wore the lower-limb exoskeleton and the signals of IMU sensors and the plantar force sensors were measured simultaneously with the walking assistance.

To compare the control timing that calculated from the proposed method and the desired timing of wearers, we also used two switches.
During the experiment, subjects held the switches and pushed them when they would like to be assisted.
When the switches were pushed, the thighs of wear will be rotated by the motors.
The timing was also simultaneously recorded with other signals.

We asked every subject to walk 10 trails.
The walking pattern were the same as the previous experiment.
So, in eight trials, subjects walked from 2 to 8 steps by starting with the left or right leg.
In the other two trails, the subjects walked and paused with random numbers of steps as they desired.
Based on the measured data, the model was also trained using the data from eight trails and tested with the data from two trails.
The prediction error of the plantar force and the assistance timing was evaluated.
The assistance timing was also compared with the switch timing.

Then the models of some subjects was copied to the small computer in the exoskeleton, the Raspberry PI 3B.
The exoskeleton was controlled by predicting the plantar force from IMU signals in real-time, and the effectiveness of the assistance was evaluated with a questionnaire.
Considering the computing performance of a small computer in the exoskeleton, we reduced the prediction frequency to 20Hz.
When the predicted walking phase changes to swing (i.e., when the plantar force on toe $f_t$ becomes lower than 50N), a command is sent to rotate the motor 0.1s at the assisting timing.

\subsubsection*{Result}

\begin{figure*}[t]
    \centering
    \includegraphics[width=0.8\linewidth]{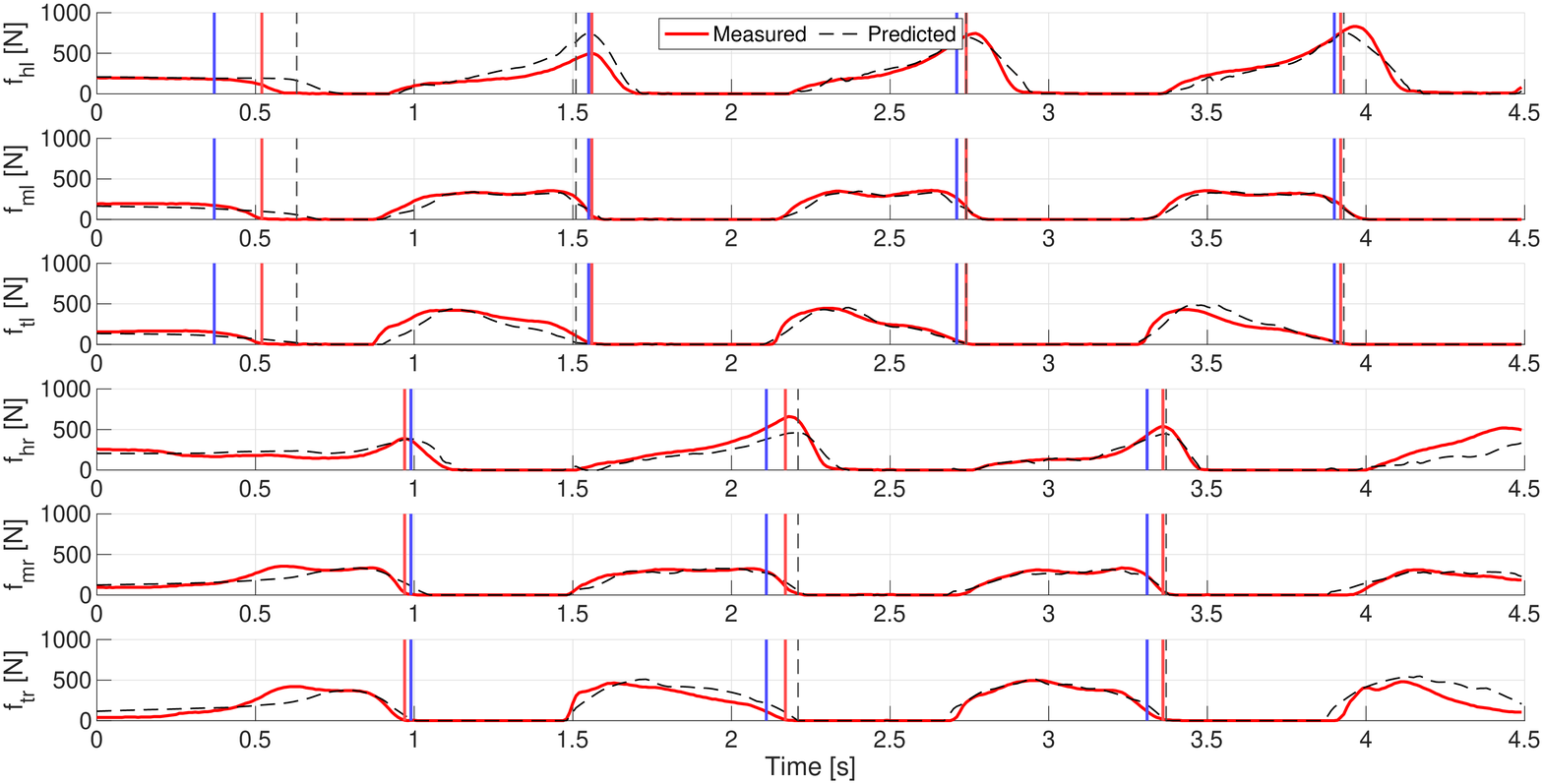}
    \caption{The prediction result for walking when exoskelton}
    \label{fig:pred_sw_dev}
\end{figure*}

In this experiment, we measured the walking data from the same nine subjects as in the previous experiment.
Fig.~\ref{fig:pred_sw_dev} shows the predicted result of a six-step walking in a trail by a subject wearing the exoskeleton and using the switches for control.
The red solid lines and the black dash lines show the predicted and measured plantar forces applied on each cell during the walking.
The same change tendency can also be predicted for both the start of walking and during the walking even with the assistance.

\begin{figure}[t]
    \centering
    \includegraphics[width=0.95\columnwidth]{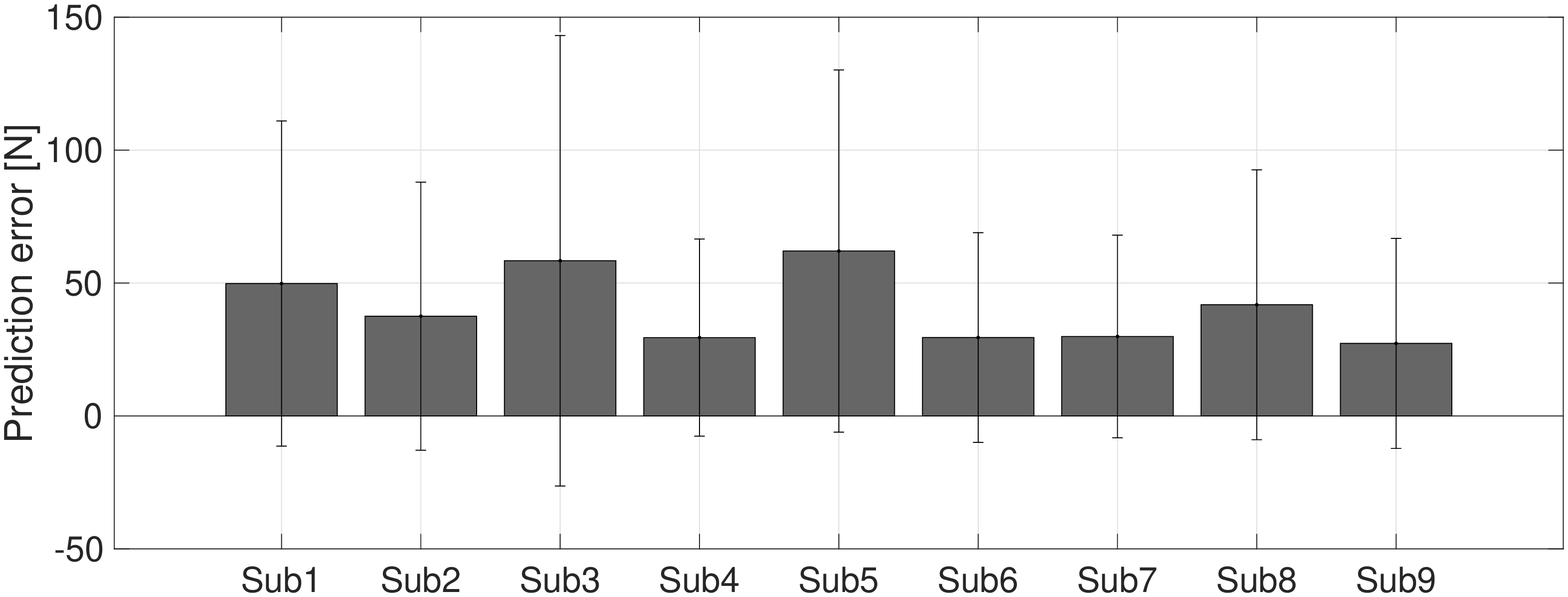}
    \caption{Prediction error for each subject when controlling the exoskelton}
    \label{fig:prederr_dev}
\end{figure}

Fig.~\ref{fig:prederr_dev} shows the average prediction errors of all subjects.
As in the previous experiment, even when wearing the exoskeleton, the plantar force can still be predicted with relatively small prediction errors (27.3\textasciitilde62.0N).
The average prediction error of all subjects is 40.3N.
Even with the assist of the exoskeleton, almost the same prediction accuracy was obtained.

The vertical lines in Fig.~\ref{fig:pred_sw_dev} show the predicted and desired control timings.
The red solid lines and the black dush lines are the timing determined from the predicted and measured plantar force.
The blue solid lines are the switched timings measured by the subjects pushing the switches in their hands.
Most of these timings are close to each other.

\begin{figure}[t]
    \centering
    \subfigure[Assistance timing while walking]{
    \includegraphics[width=0.95\columnwidth]{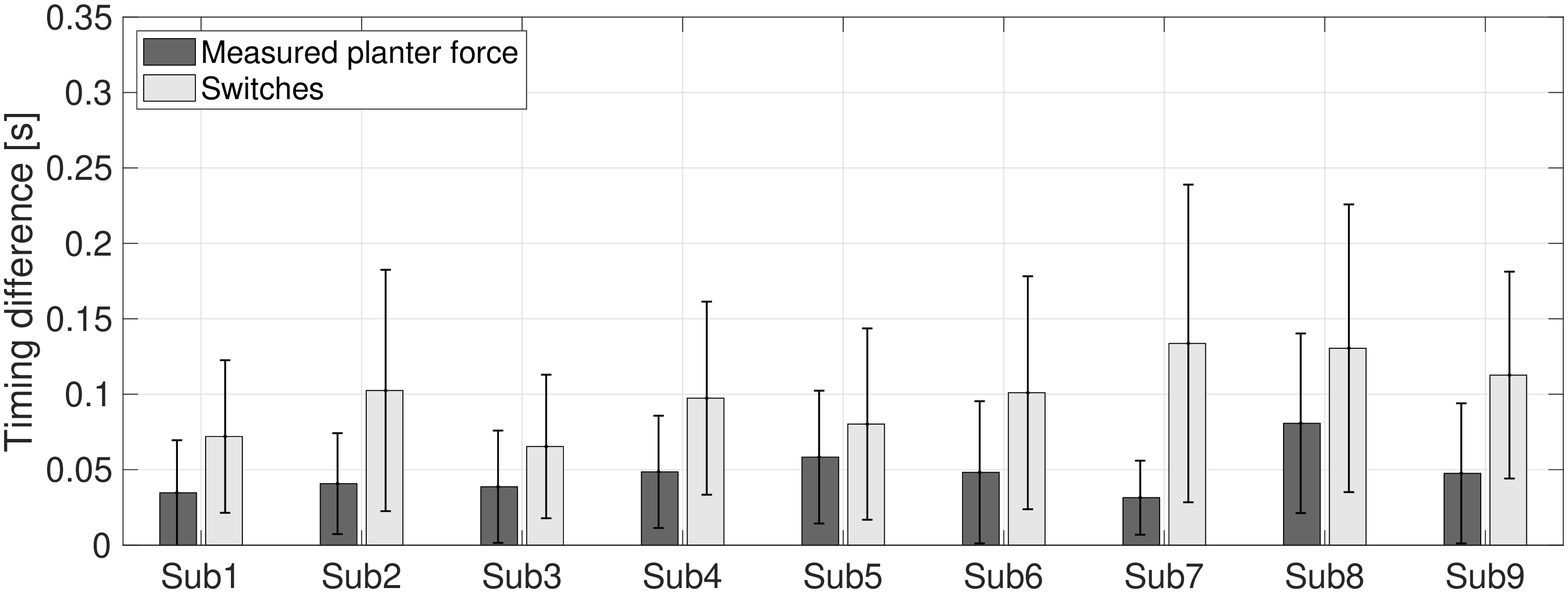}
    }
    \subfigure[Assistance timing when starting ]{
    \includegraphics[width=0.95\columnwidth]{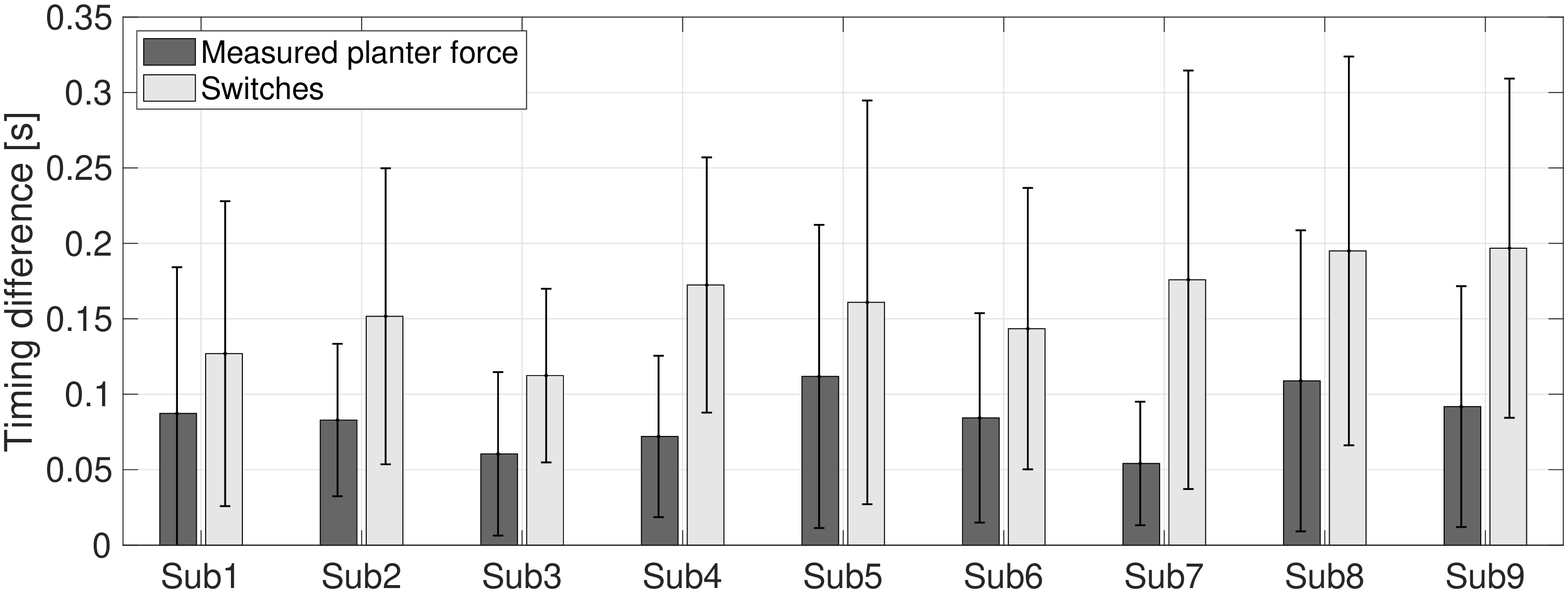}
    }
    \caption{Difference in support timing when controlling the exoskeleton}
    \label{fig:pred_tmg_err_sw}
\end{figure}

Fig.~\ref{fig:pred_tmg_err_sw} shows the time differences of the support timing between predicted timing, measured timing, and the switch timing.
The dark gray boxes show the difference between the timings calculated from the predicted and measured plantar forces.
As in previous experiment, even assisted by the exoskeleton, the proposed method can still determine the assistance timing almost as accurately as using the measured values.
As a comparison, the light gray boxes show the differences between predicted timing and the switch timing.
To assist the walking just in time, subjects should push the switches before moving their thigh due to the time-delay of the exoskeleton.
Therefore, the differences between predicted timing and switch timing are larger than compared to detecting the timing from the plantar force.
The average difference between measured and predicted timing is about 0.05s during  walking and about 0.08s when starting after having paused.

\begin{figure*}[t]
    \centering
    \includegraphics[width=0.95\linewidth]{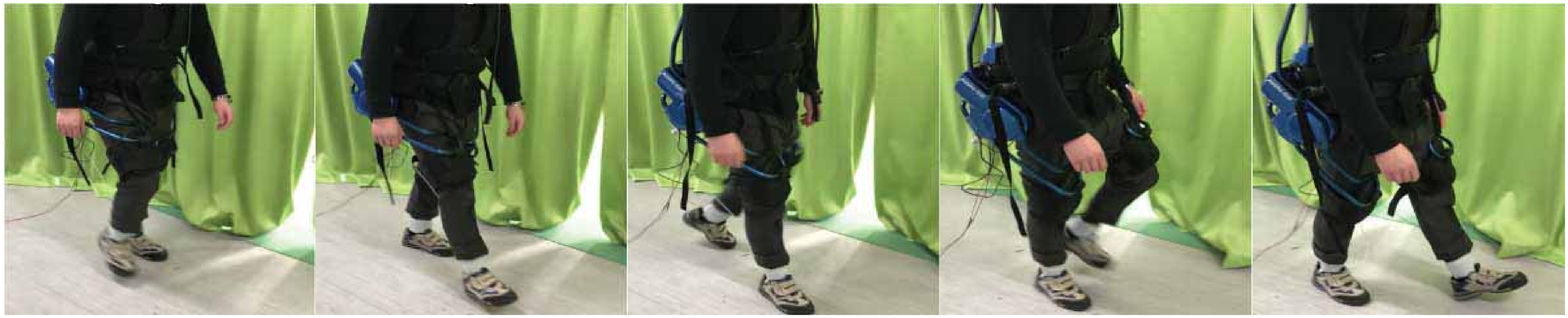}
    \caption{Walking with the assistance}
    \label{fig:walking_dev}
\end{figure*}

Finally, as show in Fig.~\ref{fig:walking_dev}, we copied the trained models of three subjects to the small computer in the exoskeleton to measure the IMU signals and control the exoskeleton in real-time.
In this case the exoskeleton was controlled by the movement of subject's leg only the measured signals of IMU sensors.
After using this exoskeleton, in questionnaires the subjects all felt that the walking motion was supported with good timing by the proposed prediction method.
This was also better than switched control.
Finally, even the start of walking was supported with good timing.

\section{Conclusion}
\label{sec:conclusion}

In this research, we proposed a method to remove the time-delay when controlling a walking assist exoskeleton by predicting the future plantar force and the walking status.
Using LSTM and fully-connected networks, the future plantar force could be predicted using only the signals measured by IMU sensors.
The predicted plantar force could be used to estimate the walking status.
Based on the estimated walking status, control commands could be sent beforehand to move the exoskeleton with the desired timing.
The experimental results showed that the predicted plantar force matches the measured value with a very small prediction error (33.9N and 40.3N when using and not using the exoskeleton).
The assistance timing can also be predicted not only during the walking but also at the start and end of walking.
The experiments also showed good performance when controlling the exoskeleton to assist the subjects in real-time.

In the future, we will gather more data from more subjects to determine whether it is possible to increase the performance of the proposed method.
We will also test the prediction accuracy when walking on different
grounds, such as slopes or stairs.

\bibliographystyle{unsrt}  
\bibliography{references}  %%% Remove comment to use the external .bib file (using bibtex).

\end{document}